\documentclass{article} 
\usepackage{iclr2027_conference,times}

\usepackage{amsmath,amsfonts,bm}

\def\eqref#1{equation~\ref{#1}}

\def\1{\bm{1}}

\DeclareMathAlphabet{\mathsfit}{\encodingdefault}{\sfdefault}{m}{sl}
\SetMathAlphabet{\mathsfit}{bold}{\encodingdefault}{\sfdefault}{bx}{n}

\usepackage{hyperref}
\usepackage{url}
\renewcommand{\cite}{\citep}
\usepackage[T1]{fontenc}    
\usepackage{hyperref}       
\usepackage{url}            
\usepackage{booktabs}       
\usepackage{amsfonts}       
\usepackage{nicefrac}       
\usepackage{microtype}      
\usepackage{xcolor}         
\definecolor{citecolor}{HTML}{0071BC}
\definecolor{linkcolor}{HTML}{D32F2F}
\definecolor{cellcolor}{HTML}{E3F2FD}
\definecolor{red}{HTML}{D32F2F}
\definecolor{magenta}{HTML}{D81B60}

\usepackage{svg}
\svgsetup{inkscapearea=page}

\usepackage{amsmath}
\usepackage{amssymb}
\usepackage{mathtools}
\usepackage{amsthm}
\usepackage{amsmath, amssymb, amsthm}
\usepackage{amsfonts}
\usepackage[capitalize,noabbrev]{cleveref}

\theoremstyle{plain}

\theoremstyle{definition}

\theoremstyle{remark}

\usepackage[textsize=tiny]{todonotes}
\usepackage{microtype}
\usepackage{graphicx}
\usepackage{booktabs} 

\usepackage{pgfplots}
\pgfplotsset{compat = newest}
\usepackage{multirow}
\usepackage{enumitem}
\usepackage{caption}
\usepackage{fancybox}
\usepackage{framed}
\usepackage{tcolorbox}

\usepackage{algorithm}
\usepackage{mathrsfs}
\usepackage{mathtools}
\usepackage{algorithmic}
\usepackage{tcolorbox}
\usepackage{listings}
\usepackage{makecell}
\usepackage{colortbl}
\usepackage{color}
\usepackage{wrapfig}
\usepackage{cancel}
\usepackage{soul,xcolor}
\usepackage{pifont}
\usepackage{tikz}
\usetikzlibrary{shapes, positioning, arrows.meta, calc, decorations.pathmorphing, quotes}
\usepackage{todonotes}
\tcbuselibrary{breakable}
\usepackage{hyperref}
\usepackage{makecell}
\usepackage{url}
\hypersetup{colorlinks=true, linkcolor=linkcolor, citecolor=citecolor,urlcolor=magenta}
\usepackage{booktabs}
\usepackage{hyperref}
\usepackage{url}
\usepackage{graphicx}
\usepackage{subcaption}
\usepackage{amsthm}

\usepackage{enumitem}
\setlist[itemize]{leftmargin=2em}
\setlist[enumerate]{leftmargin=2em}

\title{Distilled Reinforcement Learning for LLM Post-training}

\author{
Chen Wang$^{1,2}$\thanks{s-wc25@bza.edu.cn}, 
Zhaochun Li$^{2,3}$,
Jionghao Bai$^{2,4}$,
\textbf{Yining Zhang}$^{2,5}$,
\textbf{Hexuan Deng}$^{2,6}$,\\
\textbf{ Ge Lan}$^{7\dagger}$,
\textbf{Yue Wang}$^{2}$\thanks{Corresponding author}
\\[1mm]
$^{1}$ College of Elite Engineers, Nankai University, \quad
$^{2}$ Zhongguancun Academy \\
$^{3}$ Beijing Institute of Technology,\quad
$^{4}$ Zhejiang University \\
$^{5}$ Institute of Automation, Chinese Academy of Sciences \\
$^{6}$ Harbin Institute of Technology, \quad
$^{7}$ College of Software, Nankai University\\
}

\iclrfinalcopy 
\begin{document}

\maketitle

\begin{abstract}
Large language model (LLM) post-training is essential for improving reasoning, adaptation, and alignment. Existing methods mainly follow two paradigms: reinforcement learning (RL) and on-policy distillation (OPD). However, RL relies on coarse-grained outcome supervision, resulting in difficult credit assignment and limited capability to acquire new knowledge. OPD, meanwhile, unconditionally matches teacher logits through KL divergence, which creates a dilemma: similar teachers provide little new knowledge, while substantially different teachers often yield ineffective guidance, largely restricting OPD to within-family distillation. We propose Distilled Reinforcement Learning (Distilled RL), which integrates teacher supervision into the RL objective to provide fine-grained guidance, selectively transfer new knowledge and avoid unconditional imitation. Distilled RL contains three components: reverse importance sampling with clipping, negative sample reset, and sequence-level geometric normalization. Through a concise and interpretable study, we show that Distilled RL can effectively transfer knowledge from a teacher model to a student model. Extensive experiments across both within-family and cross-family (significant difference in behavior between students and teachers) distillation settings show that Distilled RL substantially outperforms baselines in terms of both pass@1 and pass@k. 

\end{abstract}

\section{Introduction}
Large language model (LLM) post-training plays a central role in improving reasoning, instruction following, and task adaptation \cite{schulman2017proximal, wei2022chain, touvron2023llama, glm2024chat, rafailov2023direct, zhong2024dpo, wang2024comprehensive, team2025kimi1_5}. Compared with supervised fine-tuning (SFT), reinforcement learning (RL) performs updates on trajectories sampled from the current policy, thereby reducing the distribution shift between training and inference and alleviating catastrophic forgetting \cite{shenfeldrl, chen2025retaining}. This on-policy training paradigm has enabled substantial advances in complex reasoning and generalization \cite{shao2024deepseekmath, liu2024deepseek, guo2024deepseek, yu2025dapo}. Inspired by the same principle, on-policy distillation (OPD) reformulates conventional knowledge distillation by replacing teacher- or dataset-generated samples with trajectories sampled from the student policy, and then minimizes the KL divergence between the student and teacher distributions on the states actually visited by the student \cite{gu2024minillm, agarwal2024policy}. In this way, OPD reduces the mismatch between distillation data and the student’s inference-time distribution while retaining dense token-level supervision from a stronger teacher \cite{yang2026learning, wu2026lightning}.

\begin{figure}[t]
\centering
\includegraphics[width=0.9\linewidth]{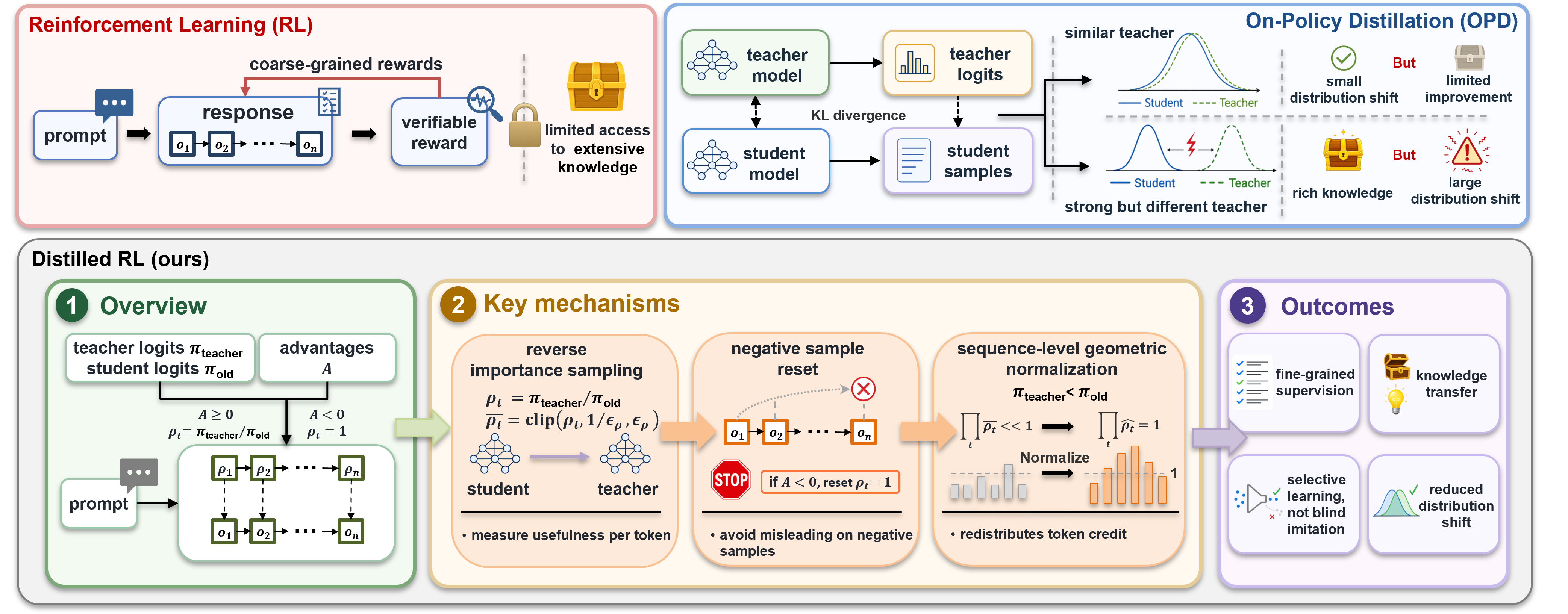}\\
\caption{Overview of Distilled RL.}
\vspace{-1em}
\label{fig:main}
\end{figure}

Despite the effectiveness of RL and OPD, both paradigms exhibit important limitations. RL usually assigns a sequence-level reward to an entire response, even though only a small subset of tokens may determine whether the final answer is correct ~\cite{shao2024deepseekmath}. Such coarse-grained supervision creates a severe credit-assignment problem and provides limited information about which intermediate reasoning steps should be encouraged or corrected \cite{lightman2023lets}. Moreover, because the learning signal is derived mainly from task rewards, RL is effective at reinforcing behaviors already accessible to the student but has limited ability to introduce knowledge that is absent from its current policy \cite{yu2025dapo, yue2025does}. OPD provides denser token-level feedback, but commonly optimizes a KL-divergence objective that unconditionally encourages the student to imitate the teacher distribution \cite{wu2025rethinking,xu2025kdrl,jin2026entropy}. This leads to a fundamental dilemma. When the teacher and student are highly similar, their distributions offer limited complementary knowledge; when they differ substantially in architecture, model family, or reasoning pattern, the resulting distribution mismatch makes token-level imitation noisy and difficult to optimize. Recent studies have similarly observed that successful OPD depends strongly on compatible teacher--student reasoning patterns and distributions \cite{wu2026lightning, fu2026revisiting, li2026rethinking}. Beyond this compatibility issue, KL-based imitation may also drive the student to align with the teacher distribution too rapidly, leading to premature convergence and leaving limited room for subsequent reward-driven improvement (see details in Subsection~\ref{subsec:obv}).

To address these limitations, we propose \textbf{Distilled Reinforcement Learning (Distilled RL)}, a unified post-training framework that incorporates teacher knowledge directly into the RL gradient (see overview in Fig.~\ref{fig:main}). Distilled RL evaluates student-generated tokens using a reverse importance ratio between the teacher policy and the old student policy, and uses this ratio to redistribute the RL learning signal at the token level. The framework contains three key components. First, \textbf{reverse importance sampling} measures the relative preference of the teacher for each student-generated token and clips the ratio to prevent unstable gradient amplification. Second, \textbf{negative sample reset} applies teacher-based reweighting only to responses with positive advantages, while resetting the importance weights of negative samples to one. This prevents the student from imitating teacher preferences along negative trajectories and preserves the original RL penalty for negative responses. Third, \textbf{sequence-level geometric normalization} normalizes the clipped token-level ratios within each response such that their geometric mean equals one. This normalization respects the multiplicative factorization of autoregressive sequence probabilities, removes systematic scale differences between teacher and student likelihoods, and retains the teacher's relative preference over tokens. Through these mechanisms, Distilled RL uses the teacher as a selective fine-grained guide rather than an unconditional imitation target. We further demonstrate through a controlled case study that Distilled RL enables the student to acquire new knowledge from the teacher that standard RL cannot effectively learn. Extensive experiments show that Distilled RL consistently outperforms RL and OPD, and, more importantly, achieves substantial gains in cross-family distillation, where conventional OPD suffers from persistent performance degradation.

Our contributions are fourfold.
\begin{itemize}
\item We identify key limitations of the two dominant post-training paradigms: RL relies on coarse-grained outcome supervision and provides limited guidance for acquiring new knowledge, while OPD performs unconditional teacher imitation, leading to rapid saturation and degraded effectiveness under substantial teacher--student distribution mismatch.
    
\item We propose Distilled RL, which integrates teacher supervision directly into the RL objective through reverse importance sampling, negative sample reset, and sequence-level geometric normalization.
    
\item We show that Distilled RL can transfer teacher-specific knowledge and effective reasoning behaviors beyond what standard outcome-reward RL can learn, supported by controlled case studies and diagnostic analyses.
    
\item Extensive experiments across within-family and cross-family settings demonstrate that Distilled RL consistently outperforms RL and OPD on reasoning, knowledge, and coding benchmarks in both pass@1 and pass@k.
\end{itemize}

\section{Related Work}
\paragraph{Reinforcement Learning for LLMs.}
Reinforcement learning has become a central approach for improving the reasoning capabilities of large language models. Early RLHF methods optimize human-preference reward models using policy-gradient algorithms such as PPO~\cite{ouyang2022training,schulman2017proximal}, while recent reinforcement learning with verifiable rewards directly evaluates responses using rule-based outcome signals. Given a prompt $q$ and a response $o$ sampled from the old policy, a general policy-optimization objective can be written as
\begin{equation}
\mathcal{J}_{\mathrm{RL}}(\theta)
=
\mathbb{E}_{\substack{
q\sim P(Q),\\
o\sim\pi_{\theta_{\mathrm{old}}}(\cdot\mid q)
}}
\left[
\frac{1}{|o|}
\sum_{t=1}^{|o|}
r_{i,t}(\theta)A(q,o)
\right],
\qquad
r_{i,t}(\theta)
=
\frac{
\pi_{\theta}
\left(
o_{i,t}\mid q,o_{i,<t}
\right)
}{
\pi_{\theta_{\mathrm{old}}}
\left(
o_{i,t}\mid q,o_{i,<t}
\right)
}.
\end{equation}
GRPO removes the need for a separate critic by estimating $A(q,o)$ from the relative rewards of a group of sampled responses~\cite{shao2024deepseekmath}, and subsequent work demonstrates that large-scale RL can elicit sophisticated reasoning behaviors from pretrained language models~\cite{guo2025deepseekr1}. Compared with supervised fine-tuning, on-policy RL trains on trajectories generated by the evolving policy, reducing the mismatch between training and inference distributions. Nevertheless, the same sequence-level advantage $A$ is generally assigned to every token in the response, even though different tokens contribute unequally to the final outcome. This coarse-grained supervision creates a challenging credit-assignment problem. Recent RL research has primarily focused on improving exploration through entropy \cite{cui2025entropy,cheng2025reasoning, wang2026targeted, wang2025arbitrary}, mitigating overthinking by controlling response length \cite{yi2025shorterbetter,liu2026length, wang2026implicit}, and providing finer-grained credit assignment via process-level rewards~\cite{luo2024improve, cui2025process, li2025treepo, cheng2026stop}. However, these methods largely optimize how existing behaviors are reinforced, rather than how new knowledge is acquired. Efficiently transferring knowledge beyond the student's current capability still requires guidance from a stronger teacher model, which is not explicitly addressed by conventional RL.

\paragraph{On-Policy Distillation.}
Knowledge distillation transfers capabilities from a stronger teacher to a smaller student by matching teacher-generated outputs or predictive distributions~\cite{hinton2015distilling,kim2016sequence}. Conventional sequence-level distillation is usually performed on fixed teacher-generated data and therefore suffers from exposure bias when the student encounters states not covered by the distillation corpus. On-policy distillation instead samples a response $o$ from the student policy and minimizes the token-level discrepancy between the student and teacher distributions on the states visited by the student:
\begin{equation}
\mathcal{L}_{\mathrm{OPD}}(\theta)
=
\mathbb{E}_{\substack{
q\sim P(Q),\\
o\sim\pi_{\theta}(\cdot\mid q)
}}
\left[
\frac{1}{|o|}
\sum_{t=1}^{|o|}
D_{\mathrm{KL}}
\left(
\pi_{\theta}(\cdot\mid q,o_{<t})
\,\Vert\,
\pi_{\text{teacher}}(\cdot\mid q,o_{<t})
\right)
\right].
\end{equation}
MiniLLM adopts reverse KL divergence and optimizes it using student-generated samples~\cite{gu2024minillm}, while generalized knowledge distillation further formalizes token-level teacher supervision along student-generated trajectories~\cite{agarwal2024policy}. Subsequent studies have explored alternative KL objectives, uncertainty-aware distillation, and combinations of knowledge distillation and reinforcement learning~\cite{wu2025rethinking,xu2025kdrl,jin2026entropy, wu2026lightning, fu2026revisiting, li2026rethinking}. However, these methods generally retain an explicit KL-based imitation objective that encourages the student to match the teacher distribution at every visited state, regardless of whether the sampled trajectory is beneficial. Such unconditional imitation can become unreliable when the teacher and student belong to different model families and exhibit substantial distributional mismatch.

\section{Method}


\subsection{Observation and Analysis \label{subsec:obv}}

\begin{figure}[t]
\centering
\includegraphics[width=1\linewidth]{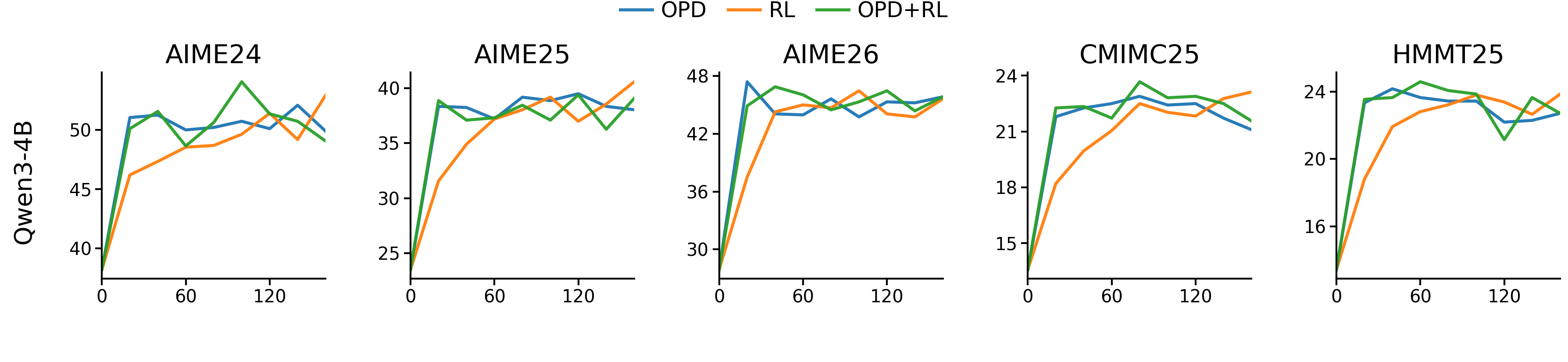}\\
\includegraphics[width=1\linewidth]{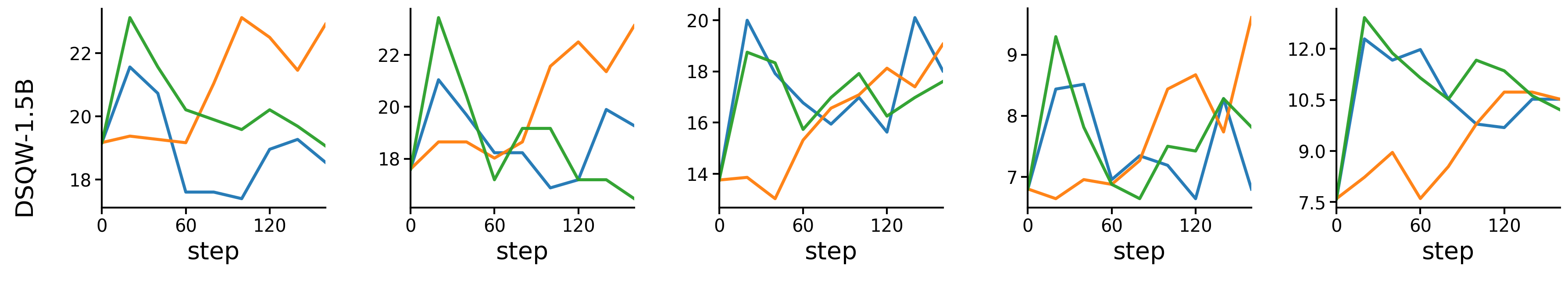}
\caption{Training dynamics of OPD, RL, and OPD+RL on five competition reasoning benchmarks. The teacher model is Qwen3-8B-GRPO, and the student models are Qwen3-4B and DeepSeek-R1-Distill-Qwen-1.5B (DSQW-1.5B). Models are trained on DAPO-17k with rollout group size $G=8$.}
\label{fig:OPD+RL}
\end{figure}

\begin{figure}[t]
\centering
\includegraphics[width=1\linewidth]{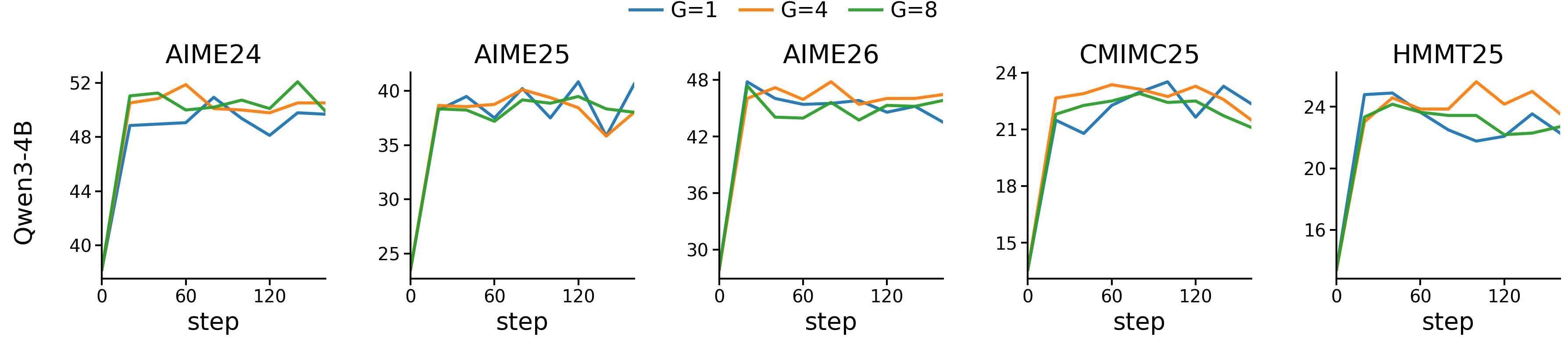}
\caption{Training dynamics of OPD with different rollout group sizes. The teacher model is Qwen3-8B-GRPO, and the student model is Qwen3-4B.}
\label{fig:OPD-G}
\end{figure}

Before presenting the formal objective of Distilled RL, we compare OPD and RL on multiple reasoning benchmarks. Figures~\ref{fig:OPD+RL} and~\ref{fig:OPD-G} show the training dynamics under different base models and rollout group sizes.

\paragraph{OPD converges rapidly but lacks sustained improvement.}
As shown in Fig.~\ref{fig:OPD+RL}, OPD improves the student model quickly at the early stage of training, but the performance soon saturates and may even decrease, especially in cross-family distillation. This suggests that KL-based imitation can rapidly align the student with the teacher on visited states, but provides limited task-aware signals for further improvement. In contrast, RL shows a slower but more sustained upward trend, since it directly optimizes task rewards through exploration. However, its sequence-level reward provides coarse-grained supervision, making learning inefficient. A natural remedy is to mix the OPD and RL losses (such as 1:1), but this strategy still exhibits a consistent performance decline in the later training stage. This may be because OPD converges rapidly and constrains the student around a KL-induced local optimum, making it difficult for the RL signal to drive sustained improvement afterward.

\paragraph{Increasing the rollout size brings limited gains to OPD.}
Fig.~\ref{fig:OPD-G} studies OPD with different rollout group sizes $G \in \{1,4,8\}$. Increasing the number of rollouts is a common variance-reduction strategy in on-policy training, as it provides a more stable estimate of relative performance. However, OPD shows similar performance across different $G$ values, and larger rollout groups do not lead to reliable or sustained gains. This indicates that the main bottleneck of OPD is not sampling variance, but its KL-based imitation objective. OPD quickly reaches a local optimum determined by teacher--student distribution matching, and additional on-policy samples are insufficient to drive continuous improvement.

These observations suggest that teacher supervision should not be introduced as an independent unconditional imitation loss. Instead, it should be coupled with the reward-driven RL objective, so that the teacher provides fine-grained token-level guidance only when the sampled response is beneficial. This motivates Distilled RL, which incorporates teacher preference directly into the RL gradient through selective and normalized importance weighting.

\subsection{Distilled RL}
Motivated by these observations, we propose \textbf{Distilled Reinforcement Learning (Distilled RL)}, which integrates teacher guidance directly into the RL objective instead of using an independent KL-based imitation loss. Given responses sampled from the old student policy, the teacher evaluates the likelihood of each student-generated token, and the resulting token-level importance weights redistribute the RL learning signal. The resulting policy optimization objective is defined as follows:
\begin{equation}
\small
\begin{aligned}
\mathcal{J}_{\text{Distilled RL}}(\theta)
=
\mathbb{E}_{\substack{
q\sim P(Q),\\
\{o_i\}_{i=1}^{G}\sim
\pi_{\theta_{\mathrm{old}}}(\cdot\mid q)
}}
\Bigg[
&\frac{1}{G}
\sum_{i=1}^{G}
\frac{1}{|o_i|}
\sum_{t=1}^{|o_i|}
\\
&\min\Bigg(
r_{i,t}(\theta)\,
w_{i,t}A_i,
\mathrm{clip}\!\left(
r_{i,t}(\theta),
1-\epsilon_{\mathrm{low}},
1+\epsilon_{\mathrm{high}}
\right)
w_{i,t}A_i
\Bigg)
\Bigg].
\end{aligned}
\end{equation}

\noindent
where
\begin{equation*}
\small
\begin{aligned}
&r_{i,t}(\theta)
=
\frac{
\pi_{\theta}
\left(
o_{i,t}\mid q,o_{i,<t}
\right)
}{
\pi_{\theta_{\mathrm{old}}}
\left(
o_{i,t}\mid q,o_{i,<t}
\right)
},\qquad
A_i
=
\frac{
R_i
-
\mathrm{mean}
\left(
\{R_j\}_{j=1}^{G}
\right)
}{
\mathrm{std}
\left(
\{R_j\}_{j=1}^{G}
\right)
},
\qquad
\rho_{i,t}
=
\frac{
\pi_{\text{teacher}}
\left(
o_{i,t}\mid q,o_{i,<t}
\right)
}{
\pi_{\theta_{\mathrm{old}}}
\left(
o_{i,t}\mid q,o_{i,<t}
\right)
}, 
\\[2mm]
&\bar{\rho}_{i,t}
=
\mathrm{clip}
\left(
\rho_{i,t},
\frac{1}{\epsilon_{\rho}},
\epsilon_{\rho}
\right),\qquad
\widetilde{\rho}_{i,t}
=
\frac{
\bar{\rho}_{i,t}
}{
\exp\left(
\frac{1}{|o_i|}
\sum_{s=1}^{|o_i|}
\log \bar{\rho}_{i,s}
\right)
},
\qquad
w_{i,t}
=
\begin{cases}
\widetilde{\rho}_{i,t},
& A_i>0,
\\
1,
& A_i\leq 0,
\end{cases}.
\end{aligned}
\end{equation*}

Here, $\pi_{\text{teacher}}$ denotes the teacher policy,
$\rho_{i,t}$ is the token-level teacher-to-student importance ratio, $\bar{\rho}_{i,t}$ is the clipped importance ratio,
and $\widetilde{\rho}_{i,t}$ is the sequence-level geometrically normalized importance weight. The effective weight $w_{i,t}$ applies the normalized ratio only to positive samples, while resetting the weight of negative samples to one. Distilled RL consists of three core components:

\paragraph{Reverse importance sampling.}
Importance sampling estimates expectations under a target distribution using samples from another distribution. In autoregressive language-model training, token-level ratios are commonly used as a practical approximation to sequence-level importance sampling, as in PPO, GRPO, and DFT \cite{schulman2017proximal, guo2025deepseek, wu2025generalization}. These methods typically perform off-policy correction by reweighting samples generated by a behavior policy toward the current policy. In Distilled RL, we reverse this direction: rather than correcting a behavior policy toward the current student policy, we reweight trajectories sampled from the current student policy toward the teacher policy. Accordingly, the teacher-to-student ratio $\rho_{i,t}$ measures how much more or less the teacher prefers each student-generated token relative to the old student policy. Although the outer policy-optimization objective already clips the student policy ratio $r_{i,t}(\theta)$, we additionally clip $\rho_{i,t}$ into $[1/\epsilon_{\rho},\epsilon_{\rho}]$ as a redundant stability safeguard against extreme teacher--student likelihood ratios.

\paragraph{Negative sample reset.}
\begin{wraptable}{r}{0.4\linewidth}
\vspace{-3mm}
\centering
\small
\caption{Effect of importance weighting.}
\label{tab:negative-reset}
\begin{tabular}{c|cc}
\hline
& $\rho_{i,t}>1$ & $\rho_{i,t}<1$ \\
\hline
$A_i>0$ & $\textcolor{blue}{\uparrow}$ & $\textcolor{blue}{\downarrow}$ \\
$A_i<0$ & $\textcolor{red}{\downarrow}$ & $\textcolor{red}{\uparrow}$ \\
\hline
\end{tabular}
\end{wraptable}
A key issue of directly applying teacher-based weights to all samples is that the effect of $\rho$ changes with the sign of the advantage.
As shown in Table~\ref{tab:negative-reset}, when $A_i>0$, teacher reweighting behaves as desired: teacher-preferred tokens with $\rho_{i,t}>1$ receive a stronger reward, while less preferred tokens with $\rho_{i,t}<1$ are relatively suppressed. This provides fine-grained guidance and enables knowledge transfer on successful trajectories. The problem appears when $A_i<0$. In this case, a teacher-preferred token receives a larger penalty, pushing the student away from the teacher. If the teacher can solve the problem, the student should imitate rather than avoid the teacher's preference; if the teacher also fails, the distillation signal is not useful. Therefore, for negative-advantage samples, we reset the importance weight to one and reduce the update to the original RL objective. This prevents teacher guidance from acting in the wrong direction and makes distillation selective. Importantly, applying teacher reweighting only to positive trajectories does not mean that Distilled RL cannot improve previously failed problems. Negative trajectories are still optimized by the original RL objective, while teacher guidance on successful trajectories can transfer reusable reasoning patterns across problems. \textbf{Prior work suggests that OPD gains largely come from imitating effective teacher reasoning paths rather than merely matching final rewards}~\cite{wang2026demystifying}. Thus, negative sample reset preserves useful reasoning-path transfer while avoiding misleading teacher reweighting on negative trajectories.

\paragraph{Sequence-level geometric normalization.}
In on-policy training, responses are sampled from the student policy rather than the teacher policy. Consequently, especially under teacher--student distribution mismatch, the teacher may assign lower probabilities than the student to many student-generated tokens, making the teacher-to-student ratios $\rho_{i,t}$ frequently smaller than one and inducing a response-specific negative offset in log-ratio space (see Appendix~\ref{app:sub_rho} for the corresponding statistics). To remove this systematic mismatch while preserving the teacher's relative preferences across tokens, Distilled RL centers the clipped log density ratios within each response:
$
\widetilde{\rho}_{i,t}
=
\frac{\bar{\rho}_{i,t}}
{\exp\!\left(
\frac{1}{|o_i|}
\sum_{s=1}^{|o_i|}
\log \bar{\rho}_{i,s}
\right)}.
$
This normalization ensures
$
\left(
\prod_{t=1}^{|o_i|}
\widetilde{\rho}_{i,t}
\right)^{1/|o_i|}
= 1,
$
while preserving pairwise relative preferences,
$
\frac{\widetilde{\rho}_{i,t}}
{\widetilde{\rho}_{i,s}}
=
\frac{\bar{\rho}_{i,t}}
{\bar{\rho}_{i,s}}.
$
Therefore, $\widetilde{\rho}_{i,t}$ measures the teacher's relative preference for each token after removing the response-level average teacher--student mismatch, yielding a within-response centered log density-ratio weighting for token-level credit redistribution.

\section{Experiments}
To validate Distilled RL, we implement it within the EasyR1 and VeRL frameworks
\cite{zheng2025easyr1, sheng2025hybridflow} and compare it against representative baselines, including OPD, RL (GRPO)~\cite{shao2024deepseekmath}, OPD+RL, DAPO~\cite{yu2025dapo}, and ExOPD~\cite{yang2026learning}. Experiments are conducted on DeepSeek-R1-Distill-Qwen-1.5B (DSQW-1.5B), Qwen3-4B, and Qwen3-1.7B~\cite{yang2024qwen2, yang2025qwen3, guo2025deepseekr1},
using DAPO-17K~\cite{yu2025dapo} for training. Teacher models include Qwen3-8B-GRPO and OpenReasoning-Nemotron-7B-GRPO \cite{ahmad2025opencodereasoning, ahmad2025opencodereasoning2}. Evaluation covers mathematical reasoning benchmarks, knowledge-intensive benchmarks, and code benchmarks. Detailed baselines, datasets, and implementation settings are provided in Appendix~\ref{app:detail}.

\subsection{Main results}

\begin{table*}[t]
\caption{
Pass@1 results on mathematical reasoning benchmarks with Qwen3-8B-GRPO as the teacher model. Distilled RL consistently improves performance over baselines.
}
\label{tab:main_results_pass1}
\centering
\resizebox{\textwidth}{!}{%
\renewcommand{\arraystretch}{1.20}
\begin{tabular}{
    >{\raggedright\arraybackslash}m{2.6cm}
    *{10}{>{\centering\arraybackslash}m{1.55cm}}|
    >{\centering\arraybackslash}m{1.45cm}
}
\toprule
\textbf{Method}
& \textbf{AIME24$_{\times32}$}
& \textbf{AIME25$_{\times32}$}
& \textbf{AIME26$_{\times32}$}
& \small{\textbf{CMIMC25$_{\times32}$}}
& \small{\textbf{HMMT25$_{\times32}$}}
& \textbf{AMC23$_{\times32}$}
& \textbf{GSM8K}
& \textbf{MATH500}
& \textbf{Minerva}
& \textbf{Olympiad}
& \textbf{Average} \\
\midrule

\rowcolor{gray!15}
\textit{Qwen3-8B-GRPO (Teacher)}
& 57.91
& 44.79
& 49.06
& 26.09
& 25.62
& 87.65
& 96.21
& 92.80
& 49.63
& 60.00
& 58.98 \\

\midrule

\rowcolor{gray!10}
Qwen3-4B
& 38.23
& 23.54
& 27.92
& 13.59
& 13.44
& 72.58
& 94.47
& 86.20
& 43.38
& 49.93
& 46.33 \\

OPD
& 52.08
& 39.47
& 45.83
& 22.50
& 23.64
& 84.68
& 94.69
& 91.80
& 46.32
& 58.66
& 55.97 \\

RL (GRPO)
& 53.64
& 42.18
& 46.45
& 23.35
& 27.39
& 87.50
& 94.69
& 91.40
& 47.42
& 60.00
& 57.40 \\

OPD+RL
& 54.06
& 39.37
& 46.45
& 22.89
& 23.85
& 84.06
& 94.84
& 92.00
& 47.05
& 59.25
& 56.38 \\

DAPO
& 51.56 & \textbf{43.65} & 45.00 & \textbf{27.03} & 23.13 & 87.58 & 94.54 & 92.60 & 46.69 & 59.56 & 57.13 
 \\

ExOPD
& 51.88
& 41.25
& 47.71
& 24.06
& 24.58
& 84.38
& 94.54
& 92.00
& 46.32
& 58.96
& 56.57 \\

\rowcolor{blue!5}
\textbf{Distilled RL}
& \textbf{56.45}
& 42.29
& \textbf{49.06}
& 26.95
& \textbf{28.85}
& \textbf{87.89}
& \textbf{94.99}
& \textbf{93.20}
& \textbf{48.89}
& \textbf{61.03}
& \textbf{58.96} \\

\rowcolor{blue!5}
\quad\textbf{$\Delta$ vs. OPD}
& \textcolor{green!70!black}{\textbf{(+4.37)}}
& \textcolor{green!70!black}{\textbf{(+2.82)}}
& \textcolor{green!70!black}{\textbf{(+3.23)}}
& \textcolor{green!70!black}{\textbf{(+4.45)}}
& \textcolor{green!70!black}{\textbf{(+5.21)}}
& \textcolor{green!70!black}{\textbf{(+3.21)}}
& \textcolor{green!70!black}{\textbf{(+0.30)}}
& \textcolor{green!70!black}{\textbf{(+1.40)}}
& \textcolor{green!70!black}{\textbf{(+2.57)}}
& \textcolor{green!70!black}{\textbf{(+2.37)}}
& \textcolor{green!70!black}{\textbf{(+2.99)}} \\

\midrule

\rowcolor{gray!10}
Qwen3-1.7B
& 22.60
& 21.35
& 18.85
& 9.14
& 11.67
& 64.30
& 88.78
& 81.80
& 35.66
& 44.44
& 39.86 \\

OPD
& 31.56
& 26.04
& 26.25
& 15.15
& 15.31
& 70.31
& 90.06
& 86.20
& 37.86
& 49.33
& 44.81 \\

RL (GRPO)
& 29.79
& 26.87
& 27.18
& 14.21
& 14.68
& 69.21
& 89.76
& 86.00
& \textbf{40.44}
& 49.48
& 44.76 \\

OPD+RL
& 30.72
& 24.58
& 26.66
& 16.09
& 15.20
& 71.64
& 90.14
& 86.20
& 38.23
& 49.48
& 44.89 \\

DAPO
& 31.56
& 25.83
& 26.35
& 16.88
& \textbf{15.52}
& 70.31
& 89.99
& 86.40
& 37.13
& 51.11
& 45.10 \\

ExOPD
& 29.79
& 26.04
& 26.88
& 15.78
& 15.00
& 71.17
& \textbf{90.22}
& 86.60
& 38.60
& 50.22
& 45.03 \\

\rowcolor{blue!5}
\textbf{Distilled RL}
& \textbf{32.70}
& \textbf{27.70}
& \textbf{28.13}
& \textbf{17.89}
& 15.20
& \textbf{71.79}
& 90.21
& \textbf{88.80}
& \textbf{40.44}
& \textbf{52.14}
& \textbf{46.50} \\

\rowcolor{blue!5}
\quad\textbf{$\Delta$ vs. OPD}
& \textcolor{green!70!black}{\textbf{(+1.14)}}
& \textcolor{green!70!black}{\textbf{(+1.66)}}
& \textcolor{green!70!black}{\textbf{(+1.88)}}
& \textcolor{green!70!black}{\textbf{(+2.74)}}
& \textcolor{red!70!black}{\textbf{(-0.11)}}
& \textcolor{green!70!black}{\textbf{(+1.48)}}
& \textcolor{green!70!black}{\textbf{(+0.15)}}
& \textcolor{green!70!black}{\textbf{(+2.60)}}
& \textcolor{green!70!black}{\textbf{(+2.58)}}
& \textcolor{green!70!black}{\textbf{(+2.81)}}
& \textcolor{green!70!black}{\textbf{(+1.69)}} \\

\midrule

\rowcolor{gray!10}
DSQW-1.5B
& 19.17
& 17.60
& 13.75
& 6.80
& 7.60
& 52.19
& 75.06
& 67.60
& 23.90
& 33.33
& 31.70 \\

OPD
& 20.72
& 19.89
& 20.10
& 8.51
& 11.97
& 57.57
& 76.64
& 73.60
& 26.10
& 37.62
& 35.27 \\

RL (GRPO)
& 21.04
& 22.70
& 18.02
& 8.82
& 10.93
& 60.62
& 80.13
& 79.80
& 27.57
& 38.96
& 36.86 \\

OPD+RL
& 21.56
& 20.41
& 18.33
& 8.28
& 11.67
& 58.59
& 81.50
& 78.40
& 25.73
& 40.88
& 36.54 \\

DAPO
& 22.29
& 21.67
& 18.02
& 9.14
& 10.83
& 59.06
& 79.61
& 76.00
& 25.37
& 37.78
& 35.98 \\

ExOPD
& 24.16
& 21.98
& 20.73
& \textbf{11.25}
& 13.54
& 65.70
& 73.69
& 78.60
& 30.14
& 35.18
& 37.50 \\

\rowcolor{blue!5}
\textbf{Distilled RL}
& \textbf{25.31}
& \textbf{24.58}
& \textbf{21.56}
& 10.23
& \textbf{14.27}
& \textbf{67.26}
& \textbf{81.57}
& \textbf{81.00}
& \textbf{31.25}
& \textbf{42.96}
& \textbf{40.00} \\

\rowcolor{blue!5}
\quad\textbf{$\Delta$ vs. OPD}
& \textcolor{green!70!black}{\textbf{(+4.59)}}
& \textcolor{green!70!black}{\textbf{(+4.69)}}
& \textcolor{green!70!black}{\textbf{(+1.46)}}
& \textcolor{green!70!black}{\textbf{(+1.72)}}
& \textcolor{green!70!black}{\textbf{(+2.30)}}
& \textcolor{green!70!black}{\textbf{(+9.69)}}
& \textcolor{green!70!black}{\textbf{(+4.93)}}
& \textcolor{green!70!black}{\textbf{(+7.40)}}
& \textcolor{green!70!black}{\textbf{(+5.15)}}
& \textcolor{green!70!black}{\textbf{(+5.34)}}
& \textcolor{green!70!black}{\textbf{(+4.73)}} \\

\bottomrule
\end{tabular}%
}
\end{table*}

\begin{figure}[t]
\centering
\includegraphics[width=0.9\linewidth]{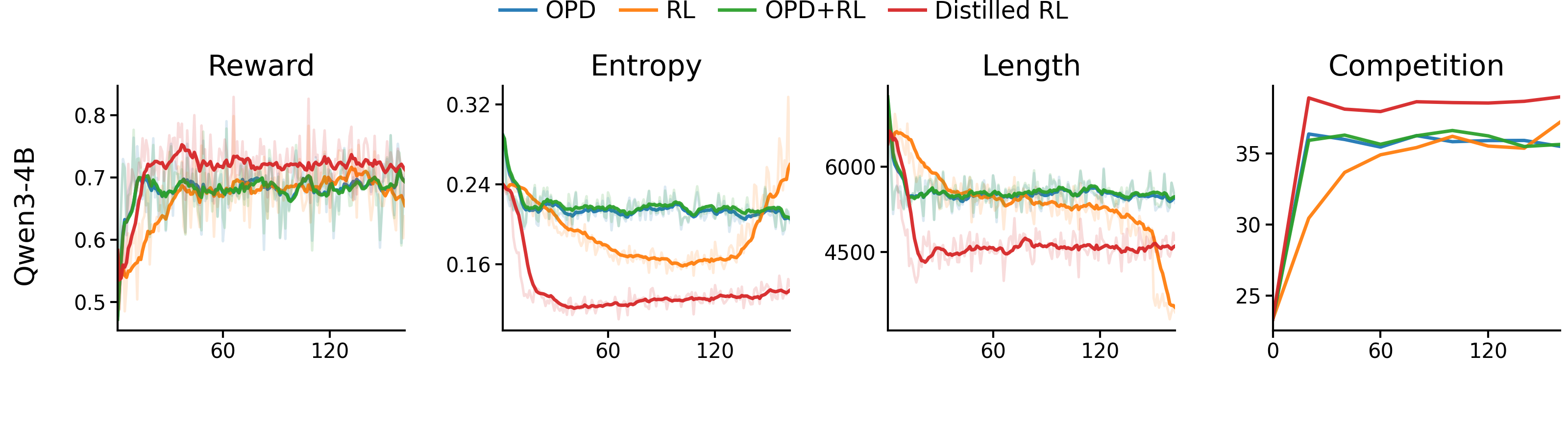}\\
\includegraphics[width=0.9\linewidth]{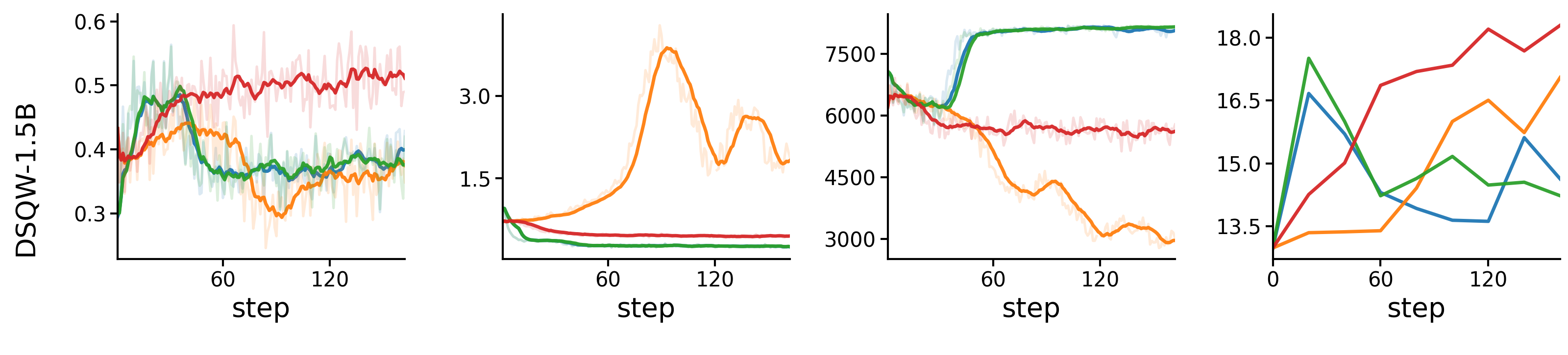}
\caption{Training dynamics of Distilled RL and baseline methods. The panels show reward, policy entropy, response length, and average Pass@1 on the five competition benchmarks in Fig.~\ref{fig:OPD+RL}.}
\label{fig:training_dynamic}
\end{figure}


\paragraph{Performance.}
As shown in Table~\ref{tab:main_results_pass1} and \ref{tab:nemotron_dsqw}, Distilled RL consistently achieves the best Pass@1 performance across all three student models. It improves the average score from 46.33 to 58.96 on Qwen3-4B, from 39.86 to 46.50 on Qwen3-1.7B, and from 31.70 to 40.00 on DSQW-1.5B, outperforming OPD by 2.99, 1.69, and 4.73 points, respectively. Moreover, using Nemotron-7B instead of Qwen3-8B as the teacher for DSQW-1.5B yields a comparable average score of 40.30, demonstrating robustness across different teachers. These results show that incorporating teacher preferences directly into the RL objective consistently improves upon standalone RL and KL-based distillation.

\paragraph{Training dynamics.}
Fig.~\ref{fig:training_dynamic} compares the optimization trajectories of different methods. Distilled RL consistently maintains higher rewards across training. Distilled RL also maintains a moderate and stable policy entropy under both model architectures, avoiding both excessive uncertainty and premature policy collapse. Furthermore, Distilled RL exhibits a stable response length throughout training. Most importantly, the average Pass@1 of Distilled RL continues to improve steadily over the course of training. In contrast, RL improves slowly, OPD converges prematurely after rapid early progress, and OPD+RL provides no clear additional gain. These observations support the motivation of Distilled RL: rather than introducing a separate imitation objective, teacher supervision directly redistributes the RL learning signal, resulting in more stable optimization and sustained performance improvement.

\paragraph{Generalization.}
Beyond the mathematical reasoning benchmarks used for training evaluation, Table~\ref{tab:additional_results} further reports Pass@1 results on knowledge-intensive and coding benchmarks, together with Pass@16 results on mathematical reasoning benchmarks. Distilled RL achieves strong performance on MMLU-Pro and SuperGPQA, indicating that the proposed objective preserves general knowledge capabilities during post-training. On LiveCodeBench, Distilled RL further improves the Qwen3-4B student from 44.45 to 56.02, outperforming OPD, RL (GRPO), and OPD+RL by 2.94, 3.22, and 3.79 points, respectively. This result suggests that the benefits of Distilled RL can generalize beyond the mathematical domain used for post-training to code generation and programming reasoning. Under Pass@16 evaluation, Distilled RL also achieves the strongest or competitive performance across most competition benchmarks, suggesting that it improves not only the quality of the highest-probability response but also the overall response distribution.

\paragraph{Cross-family distillation.}
The advantage of Distilled RL is evident in the cross-family setting. For DSQW-1.5B, OPD improves over the base model but remains consistently behind RL, suggesting that direct KL-based imitation becomes less effective when the teacher and student exhibit substantially different reasoning distributions. Simply combining OPD with RL also fails to consistently outperform RL, indicating that unconditional teacher imitation may interfere with reward optimization. In contrast, Distilled RL achieves the largest improvement over all baselines. These results suggest that selectively incorporating teacher preferences into the RL objective is substantially more robust than explicit distribution matching, especially when the teacher and student belong to different model families.

Distilled RL provides a more effective way to combine reinforcement learning and teacher supervision. By replacing unconditional KL imitation with selective importance-weighted guidance, it consistently improves Pass@1 and Pass@k across different student models, with especially clear gains in cross-family distillation. In addition, Appendix~\ref{app:more_results} further reports compute-cost analysis, distillation with an alternative teacher, additional Qwen3-1.7B generalization results, failure cases with overly compressed teachers, and detailed teacher--student importance-ratio statistics.

\subsection{Teacher Information Transfer}
\begin{wrapfigure}{r}{0.3\linewidth}
\vspace{-4mm}
\centering
\includegraphics[width=\linewidth]{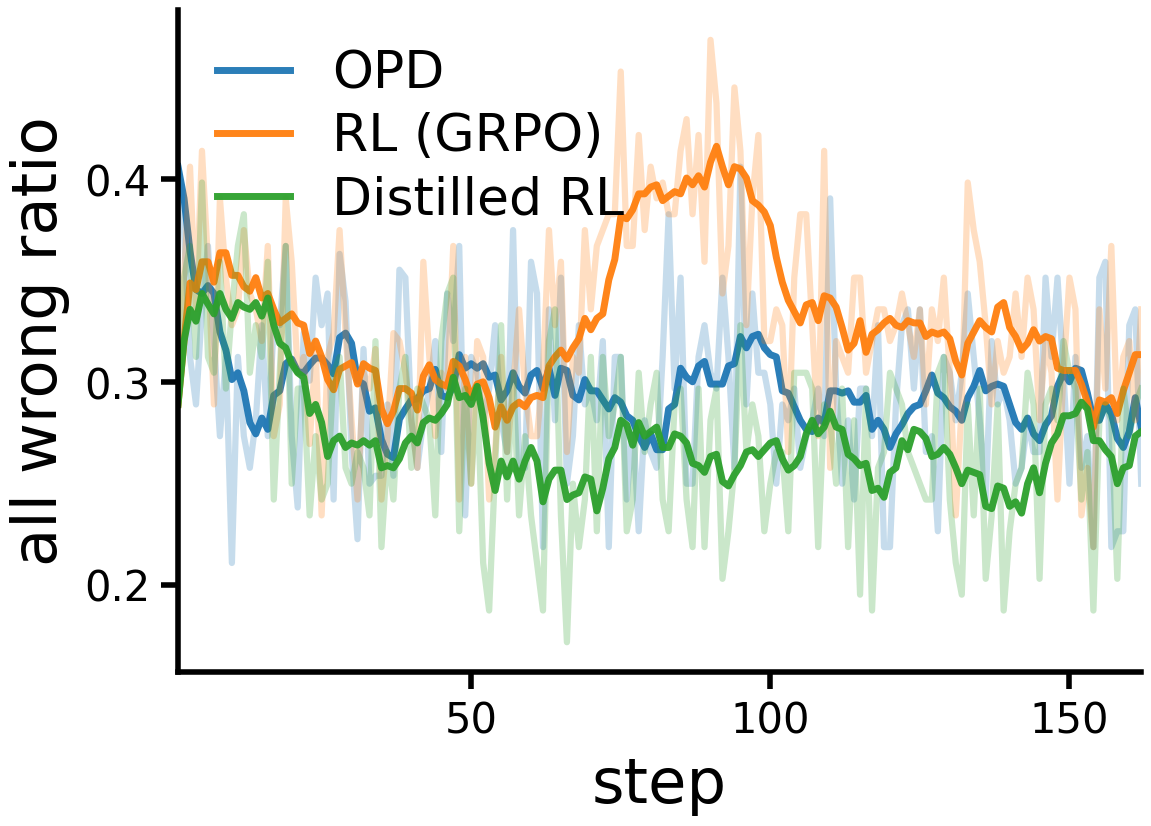}
\caption{All-wrong ratio during training.}
\label{fig:wrong-case}
\vspace{-4mm}
\end{wrapfigure}

\paragraph{Generalization from teacher knowledge.}
Similar to OPD, Distilled RL learns from the teacher through the transfer of effective reasoning paths. To examine this, we track the \emph{all-wrong ratio}, i.e., the fraction of prompts for which all sampled responses are incorrect, during training. As shown in Fig.~\ref{fig:wrong-case}, Distilled RL rapidly reduces the all-wrong ratio and maintains a low level compared with OPD and RL. This indicates that learning teacher-preferred reasoning paths from successful trajectories can generalize beyond those trajectories and help the student solve problems on which it previously failed. Thus, negative sample reset does not prevent knowledge acquisition on difficult problems.

\paragraph{Interpretable entropy case.}
\begin{wrapfigure}{r}{0.3\linewidth}
\vspace{-4mm}
\centering
\includegraphics[width=\linewidth]{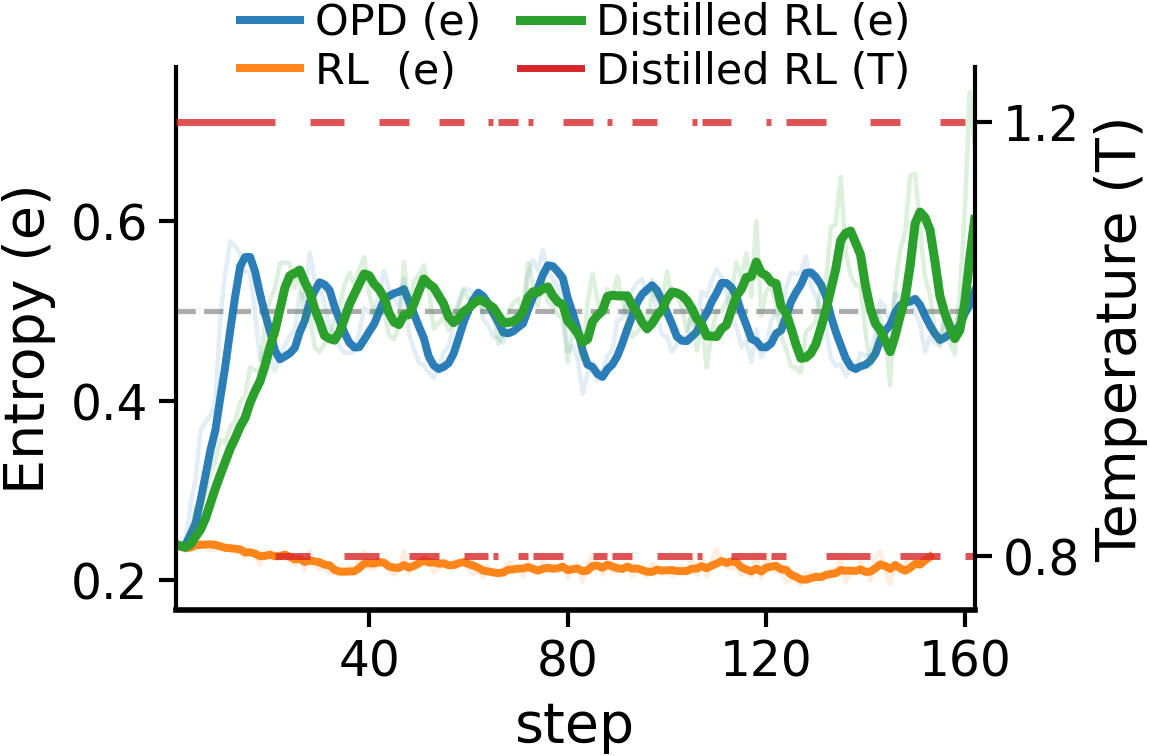}
\caption{Entropy trajectories under a temperature-controlled teacher.}
\label{fig:entropy-case}
\vspace{-4mm}
\end{wrapfigure}

We further design a controlled entropy experiment to test whether Distilled RL can acquire teacher-side distributional properties beyond the outcome reward. Entropy is a convenient signal for this purpose because it can be directly controlled through the teacher temperature and is not contained in the task reward. To isolate teacher-side information from differences in model scale or architecture, we construct the teacher from the student model itself and vary only its temperature: we use temperature $0.8$ when the student entropy exceeds $0.5$, and $1.2$ otherwise. Therefore, if the student successfully acquires the teacher-side information, its entropy should fluctuate around $0.5$. As shown in Fig.~\ref{fig:entropy-case}, Distilled RL and OPD adapt their entropy toward the target region in response to the teacher temperature, whereas RL does not exhibit the same tracking behavior. This provides an interpretable demonstration that Distilled RL can transfer teacher-side distributional properties through the policy-gradient objective.

\begin{table*}[t]
\caption{
Additional Pass@1 and Pass@16 results. Distilled RL achieves strong generalization performance across knowledge-intensive, coding, and mathematical reasoning benchmarks. “-” indicates that no successful solution was obtained across all evaluated samples.
}
\small
\label{tab:additional_results}
\centering
\resizebox{\textwidth}{!}{%
\renewcommand{\arraystretch}{1.35}
\begin{tabular}{
    >{\raggedright\arraybackslash}m{2.8cm}
    *{2}{>{\centering\arraybackslash}m{1.7cm}}
    *{1}{>{\centering\arraybackslash}m{2.1cm}}
    *{6}{>{\centering\arraybackslash}m{1.55cm}}
}
\toprule
\multirow{2}{*}{\textbf{Methods}}
& \multicolumn{3}{c}{\textbf{Pass@1}}
& \multicolumn{6}{c}{\textbf{Pass@16}} \\
\cmidrule(lr){2-4}
\cmidrule(lr){5-10}
& \textbf{MMLU$_{\text{pro}}$}
& \textbf{SuperGPQA}
& \textbf{LiveCodeBench}
& \textbf{AIME24}
& \textbf{AIME25}
& \textbf{AIME26}
& \textbf{AMC23}
& \textbf{CMIMC25}
& \textbf{HMMT25} \\
\midrule

\rowcolor{gray!10}
Qwen3-4B
& 72.86
& 32.60
& 44.45
& 68.33
& 43.33
& 45.00
& 87.50
& 30.00
& 35.00 \\

OPD
& 73.03
& 38.40
& 53.08
& 73.33
& 58.33
& 65.00
& 95.00
& 40.00
& 41.67 \\

RL (GRPO)
& 72.86
& 38.60
& 52.80
& \textbf{80.00}
& 65.00
& 68.33
& \textbf{97.50}
& 38.75
& \textbf{50.00} \\

OPD+RL
& 73.21
& 38.00
& 52.23
& 75.00
& 56.67
& 68.33
& 93.75
& 40.00
& 41.67 \\

\rowcolor{blue!5}
\textbf{Distilled RL}
& \textbf{74.46}
& \textbf{39.60}
& \textbf{56.02}
& \textbf{80.00}
& \textbf{66.67}
& \textbf{70.00}
& \textbf{97.50}
& \textbf{47.50}
& 46.67 \\

\midrule

\rowcolor{gray!10}
DSQW-1.5B
& 29.46
& 16.00
& -
& 58.33
& 41.67
& 40.00
& 87.50
& 20.00
& 21.67 \\

OPD
& 37.50
& 21.00
& -
& 55.00
& \textbf{43.33}
& 43.33
& 93.75
& 23.75
& \textbf{31.67} \\

RL (GRPO)
& 32.67
& 20.60
& -
& 60.00
& 38.33
& 43.33
& 92.50
& 22.50
& 26.67 \\

OPD+RL
& \textbf{37.85}
& 22.00
& -
& 58.33
& 41.67
& 48.33
& \textbf{95.00}
& 22.50
& 26.67 \\

\rowcolor{blue!5}
\textbf{Distilled RL}
& 36.25
& \textbf{22.20}
& -
& \textbf{70.00}
& \textbf{43.33}
& \textbf{50.00}
& \textbf{95.00}
& \textbf{27.50}
& \textbf{31.67} \\

\bottomrule
\end{tabular}%
}
\end{table*}

\begin{table*}[t]
\small
\caption{Ablation results of Distilled RL on mathematical reasoning and knowledge-intensive benchmarks. Values in parentheses denote the performance difference relative to the complete Distilled RL method.}
\label{tab:ablation}
\centering
\resizebox{\textwidth}{!}{%
\renewcommand{\arraystretch}{1.30}
\begin{tabular}{
    >{\raggedright\arraybackslash}m{3.0cm}
    *{5}{>{\centering\arraybackslash}m{1.5cm}}
    >{\centering\arraybackslash}m{1.5cm}
    >{\centering\arraybackslash}m{1.5cm}
    >{\centering\arraybackslash}m{1.7cm}
}
\toprule
\multirow{2}{*}{\textbf{Method}}
& \textbf{AIME24}
& \textbf{AIME25}
& \textbf{AIME26}
& \textbf{CMIMC25}
& \textbf{HMMT25}
& \multirow{2}{*}{\textbf{AVG}}
& \multirow{2}{*}{\textbf{MMLU$_{\text{pro}}$}}
& \multirow{2}{*}{\textbf{SuperGPQA}} \\
& \textbf{AMC23}
& \textbf{GSM8K}
& \textbf{MATH500}
& \textbf{Minerva}
& \textbf{Olympiad}
& & & \\
\midrule
\multicolumn{9}{c}{\textbf{DSQW-1.5B}} \\
\midrule

\multirow{2}{*}{w/o Negative Reset}
& 19.48 & 17.71 & 14.06 & 7.11 & 8.85
& 33.61 & 34.10 & 18.80 \\
& 54.53 & 79.30 & 73.00 & 27.20 & 34.81
& \textcolor{red}{(-6.39)}
& \textcolor{red}{(-2.15)}
& \textcolor{red}{(-3.40)} \\

\multirow{2}{*}{w/o Geometric Norm.}
& 24.37 & 23.22 & 19.79 & 9.21 & 13.33
& 38.76 & 33.57 & 21.60 \\
& 66.01 & 80.51 & 80.80 & 29.04 & 41.33
& \textcolor{red}{(-1.24)}
& \textcolor{red}{(-2.68)}
& \textcolor{red}{(-0.60)} \\
\midrule
\multicolumn{9}{c}{\textbf{Qwen3-4B}} \\
\midrule

\multirow{2}{*}{w/o Negative Reset}
& 38.44 & 33.85 & 39.48 & 17.89 & 19.38
& 50.15 & 71.43 & 34.80 \\
& 79.53 & 92.49 & 88.40 & 41.54 & 50.52
& \textcolor{red}{(-8.81)}
& \textcolor{red}{(-3.03)}
& \textcolor{red}{(-4.80)} \\

\multirow{2}{*}{w/o Geometric Norm.}
& 54.08 & 41.73 & 48.54 & 24.06 & 26.35
& 57.49 & 72.50 & 37.40 \\
& 86.56 & 94.54 & 92.60 & 46.69 & 59.70
& \textcolor{red}{(-1.47)}
& \textcolor{red}{(-1.96)}
& \textcolor{red}{(-2.20)} \\

\bottomrule
\end{tabular}
}
\end{table*}

\subsection{Ablation Study}

\begin{wrapfigure}{r}{0.35\linewidth}
\vspace{-4mm}
\centering
\includegraphics[width=\linewidth]{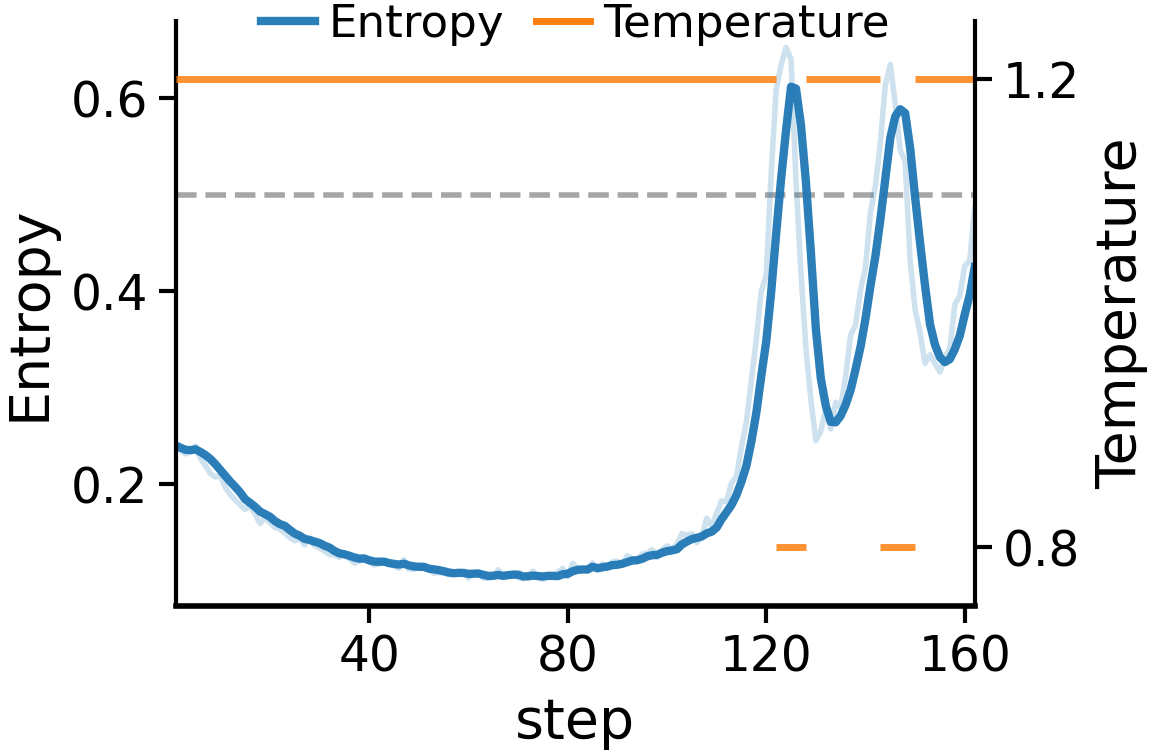}
\caption{Entropy trajectory without negative sample reset. 
}
\label{fig:ab-entropy-case}
\vspace{-4mm}
\end{wrapfigure}

\paragraph{Negative sample reset.}
As shown in Table~\ref{tab:ablation}, removing negative sample reset consistently causes the largest performance degradation on both Qwen3-4B and DSQW-1.5B. The average Pass@1 decreases by 8.81 and 6.39 points, respectively, indicating that teacher supervision on negative-advantage trajectories is harmful rather than beneficial. This observation agrees with our analysis in Table~\ref{tab:negative-reset}: when a sampled response receives a negative advantage, teacher-preferred tokens are assigned even larger penalties, forcing the student away from the teacher distribution. Fig.~\ref{fig:ab-entropy-case} further provides an interpretable demonstration. Under the same entropy-control setting as Fig.~\ref{fig:entropy-case}, removing the negative sample reset prevents the student from following the target entropy specified by the teacher. Instead of approaching the desired entropy region, the student's entropy remains substantially below the target and fails to respond to the teacher's temperature changes. This result suggests that applying teacher guidance to negative trajectories suppresses rather than facilitates knowledge transfer, preventing the student from acquiring new information from the teacher.

\paragraph{Sequence-level geometric normalization.}
As shown in Table~\ref{tab:ablation}, removing sequence-level geometric normalization also consistently degrades performance, although to a smaller extent. Since the sampled tokens are generated by the student policy rather than the teacher policy, the teacher typically assigns lower probabilities to many student-generated tokens, making the token-level importance ratios frequently smaller than one. Consequently, the product of these ratios decreases rapidly with sequence length, systematically weakening the optimization signal for positive trajectories. Sequence-level geometric normalization removes this global scaling effect while preserving the teacher's relative preference among tokens, enabling stable token-level credit redistribution without altering the overall optimization strength.

\section{Conclusion}

We presented Distilled Reinforcement Learning (Distilled RL), a unified post-training framework that integrates teacher supervision directly into the reinforcement learning objective. Unlike conventional on-policy distillation, which relies on unconditional KL-based imitation, Distilled RL uses reverse importance sampling to selectively redistribute the RL learning signal according to teacher preferences. The proposed framework consists of three simple yet effective components: reverse importance sampling, negative sample reset, and sequence-level geometric normalization. Together, these mechanisms enable fine-grained knowledge transfer while preserving the optimization behavior of reinforcement learning. Through an interpretable entropy-control case study, we further demonstrated that Distilled RL can transfer previously unavailable knowledge from the teacher to the student beyond the original reward signal. Extensive experiments on both within-family and cross-family distillation consistently showed that Distilled RL outperforms RL, OPD, and their direct combination across mathematical reasoning, knowledge-intensive benchmarks, and Pass@k evaluation. We hope Distilled RL provides a simple and effective paradigm for combining reinforcement learning and knowledge distillation in future LLM post-training.

\section*{AI use statement}
Generative AI tools were used solely to assist with translation and linguistic polishing of author-written text, including improvements to grammar, wording, and readability. They were not used for research ideation, conceptualization, method development, experimental design or execution, implementation, data analysis, result interpretation, or any other scientific contribution. All AI-assisted edits were reviewed and verified by the authors, who take full responsibility for the final content of this work.

\bibliography{reference}
\bibliographystyle{iclr2027_conference}
\newpage
\appendix
\section{Limitation}
\addcontentsline{toc}{section}{Limitation}
A fundamental limitation of on-policy distillation is that the teacher is only queried to evaluate student-generated trajectories, without sampling complete responses from the teacher. Consequently, the training procedure cannot directly determine whether the teacher is capable of solving a given problem. Distilled RL mitigates much of the resulting risk through negative sample reset, which removes teacher-based reweighting from negative student trajectories. Nevertheless, an implicit assumption remains: for problems that the student solves correctly, the teacher should also possess sufficient competence such that the token-level preferences provide useful guidance. This assumption is theoretically reasonable because the teacher is generally expected to be stronger than the student. In practice, however, teacher superiority is neither uniform across different prompts nor guaranteed throughout the dynamic training process. As the student improves, it may become comparable to or even locally outperform the fixed teacher on certain problems, in which case teacher-based reweighting may introduce suboptimal or misleading preferences into otherwise successful trajectories.

Beyond teacher capability, the \emph{distillability} of the teacher's reasoning process is also important. As shown in Table~\ref{tab:teacher_compression}, two teachers with nearly identical accuracy can lead to noticeably different student performance when one teacher is explicitly encouraged to produce shorter reasoning trajectories. An overly concise teacher may provide token-level preferences that favor prematurely shortened reasoning, which can be difficult for a smaller student to reproduce and may suppress necessary intermediate reasoning steps. Therefore, in practice, teacher selection or teacher training should avoid models whose reasoning is excessively compressed, even when their final-answer accuracy remains strong.

Future work could alleviate these limitations by explicitly estimating teacher competence and distillability, for example, through occasional teacher-side rollouts, verifier-based correctness checks, confidence-aware gating, or adaptive attenuation of distillation weights when the teacher provides uncertain, inconsistent, or overly compressed guidance.

\section{More Distillation Results}
\label{app:more_results}

\subsection{Compute cost comparison.}
\begin{wrapfigure}{r}{0.38\linewidth}
\vspace{-4mm}
\centering
\includegraphics[width=\linewidth]{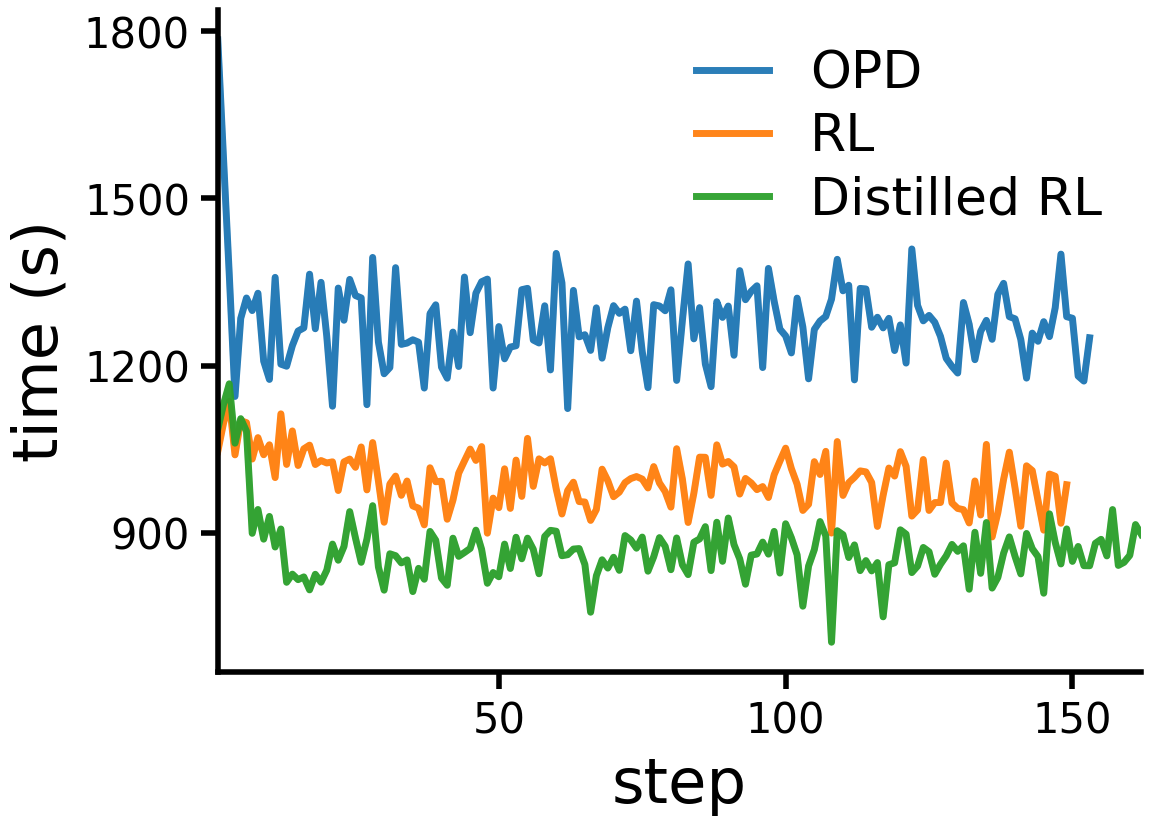}
\caption{
Per-step training time for Qwen3-4B with Qwen3-8B-GRPO as the teacher.
}
\label{fig:compute_time}
\vspace{-4mm}
\end{wrapfigure}

Table~\ref{tab:compute_cost} compares the computational requirements of different training objectives. All methods require two forward passes and one backward pass per update. RL (GRPO) evaluates $\pi_{\theta}$ and $\pi_{\mathrm{ref}}$, whereas OPD and Distilled RL evaluate $\pi_{\theta}$ and $\pi_{\mathrm{teacher}}$. Since the teacher is slightly larger than the student or reference model, OPD and Distilled RL are theoretically somewhat more expensive than RL in terms of model computation, but the additional overhead is marginal.

In practice, however, the total training cost also depends strongly on rollout length. Distilled RL rapidly learns the teacher's more concise reasoning behavior, resulting in shorter generated responses. As shown in Fig.~\ref{fig:compute_time}, its wall-clock time per training step is consistently lower than RL after the initial training stage, despite using the larger teacher model. This indicates that the reduced rollout length can more than offset the additional teacher-forward cost in practice.

\begin{table}[h]
\centering
\small
\caption{
Compute-cost comparison of different training objectives.
All methods require two forward passes and one backward pass per update.
}
\label{tab:compute_cost}
\renewcommand{\arraystretch}{1.20}
\setlength{\tabcolsep}{7pt}
\begin{tabular}{lccc}
\toprule
\textbf{Method}
& \textbf{Policy Forward}
& \textbf{Auxiliary Forward}
& \textbf{Backward} \\
\midrule
RL (GRPO)
& $\pi_{\theta_{\rm{old}}}$
& $\pi_{\mathrm{ref}}$
& $\pi_{\theta}$ \\

OPD
& $\pi_{\theta_{\rm{old}}}$
& $\pi_{\mathrm{teacher}}$
& $\pi_{\theta}$ \\

Distilled RL
& $\pi_{\theta_{\rm{old}}}$
& $\pi_{\mathrm{teacher}}$
& $\pi_{\theta}$ \\
\bottomrule
\end{tabular}
\end{table}

\subsection{Distillation with Nemotron-7B-GRPO.}
To further evaluate Distilled RL under a different teacher model, we conduct an additional distillation experiment using Nemotron-7B-GRPO as the teacher and DSQW-1.5B as the student. The results are summarized in Table~\ref{tab:nemotron_dsqw}.

\subsection{More results on Qwen3-1.7B.}
We further report the generalization performance of Qwen3-1.7B in Table~\ref{tab:qwen17_generalization}, complementing the main results reported for Qwen3-4B and DSQW-1.5B. We evaluate knowledge-intensive capabilities on MMLU-Pro and SuperGPQA, coding capability on LiveCodeBench, and multi-sampling reasoning performance using Pass@16. Distilled RL achieves the best Pass@1 performance on all three generalization benchmarks, reaching 65.89 on MMLU-Pro, 31.60 on SuperGPQA, and 38.67 on LiveCodeBench. In particular, the improvement on LiveCodeBench indicates that the benefits of Distilled RL extend beyond the mathematical domain used for post-training to code generation and programming reasoning. Distilled RL also obtains the strongest or tied-best Pass@16 performance across all six reasoning benchmarks, further demonstrating consistent improvements in the overall response distribution.

\subsection{When does Distilled RL fail?}
Distilled RL relies not only on the capability of the teacher, but also on whether its reasoning process is sufficiently informative and learnable by the student. To study this factor, we compare two Qwen3-8B teachers: a standard GRPO-trained model and a variant additionally trained with a length penalty to encourage more concise reasoning. As shown in Table~\ref{tab:teacher_compression}, the two teachers achieve nearly identical average accuracy (58.98 vs.\ 58.92), while the length-penalized teacher produces substantially shorter responses. However, Distilled RL with the compressed teacher decreases the student accuracy from 58.96 to 56.88.

\begin{table*}[t]
\caption{
Effect of teacher reasoning compression on Distilled RL. The two teachers have comparable accuracy, while the length-penalized teacher produces shorter responses and leads to weaker student performance.
}
\label{tab:teacher_compression}
\centering
\resizebox{\textwidth}{!}{%
\renewcommand{\arraystretch}{1.22}
\begin{tabular}{
>{\raggedright\arraybackslash}m{2.8cm}
*{10}{>{\centering\arraybackslash}m{1.45cm}}|
>{\centering\arraybackslash}m{1.45cm}
}
\toprule
\textbf{Setting / Metric}
& \textbf{AIME24}
& \textbf{AIME25}
& \textbf{AIME26}
& \textbf{CMIMC25}
& \textbf{HMMT25}
& \textbf{AMC23}
& \textbf{GSM8K}
& \textbf{MATH500}
& \textbf{Minerva}
& \textbf{Olympiad}
& \textbf{Average} \\
\midrule
\multicolumn{12}{c}{\textbf{Qwen3-8B-GRPO Teacher}} \\
\midrule
Teacher Accuracy
& 57.91
& 44.79
& 49.06
& 26.09
& 25.62
& 87.65
& 96.21
& 92.80
& 49.63
& 60.00
& 58.98 \\
Teacher Length
& 6079.9
& 6643.7
& 6332.4
& 7521.1
& 7363.0
& 4460.0
& 1097.4
& 2704.0
& 3272.8
& 4885.2
& 5035.90 \\
\rowcolor{blue!5}
\textbf{Distilled RL}
& \textbf{56.45}
& \textbf{42.29}
& \textbf{49.06}
& \textbf{26.95}
& \textbf{28.85}
& \textbf{87.89}
& \textbf{94.99}
& \textbf{93.20}
& \textbf{48.89}
& \textbf{61.03}
& \textbf{58.96} \\
\midrule
\multicolumn{12}{c}{
\textbf{Qwen3-8B-GRPO Teacher with Length Penalty}
} \\
\midrule
Teacher Accuracy
& 57.39
& 40.83
& 50.00
& 27.96
& 28.22
& 85.46
& 95.83
& 93.20
& 48.52
& 61.78
& 58.92 \\
Teacher Length
& 5793.8
& 6058.7
& 5690.5
& 6809.5
& 6830.2
& 3287.3
& 755.5
& 2108.1
& 2206.0
& 4019.2
& 4355.86 \\
\rowcolor{blue!5}
\textbf{Distilled RL}
& 54.48
& 40.31
& 43.85
& 27.11
& 24.58
& 87.27
& 94.31
& 91.60
& 45.59
& 59.70
& 56.88 \\
\bottomrule
\end{tabular}%
}
\end{table*}
 
This result suggests that teacher strength alone is insufficient for effective Distilled RL: the teacher must also provide reasoning trajectories that are sufficiently robust and learnable by the student. An overly compressed teacher may solve a problem correctly with a concise reasoning path that is difficult for a smaller student to reproduce. Since Distilled RL transfers the teacher's token-level preferences, such a teacher can systematically favor prematurely concise reasoning, which may suppress the student's necessary intermediate reasoning steps. Therefore, a stronger or more concise teacher is not necessarily a better teacher for distillation; the \emph{distillability} of its reasoning process is also important.

\begin{table*}[t]
\caption{
Pass@1 results for distillation from Nemotron-7B-GRPO
to DSQW-1.5B. Bold numbers indicate the best performance among student-side methods, excluding the teacher.
}
\label{tab:nemotron_dsqw}
\centering
\resizebox{\textwidth}{!}{%
\renewcommand{\arraystretch}{1.25}
\begin{tabular}{
    >{\raggedright\arraybackslash}m{2.8cm}
    *{10}{>{\centering\arraybackslash}m{1.45cm}}|
    >{\centering\arraybackslash}m{1.45cm}
}
\toprule
\textbf{Method}
& \textbf{AIME24}
& \textbf{AIME25}
& \textbf{AIME26}
& \textbf{CMIMC25}
& \textbf{HMMT25}
& \textbf{AMC23}
& \textbf{GSM8K}
& \textbf{MATH500}
& \textbf{Minerva}
& \textbf{Olympiad}
& \textbf{Average} \\
\midrule

\rowcolor{gray!15}
\textit{Nemotron-7B-GRPO (Teacher)}
& 54.79
& 50.00
& 52.81
& 27.11
& 30.63
& 85.95
& 91.74
& 89.40
& 42.28
& 59.26
& 58.40 \\
\midrule
\rowcolor{gray!10}
DSQW-1.5B
& 19.17
& 17.60
& 13.75
& 6.80
& 7.60
& 52.19
& 75.06
& 67.60
& 23.90
& 33.33
& 31.70 \\

OPD
& 20.31
& 22.40
& 18.02
& 8.91
& \textbf{12.40}
& 67.34
& 85.14
& 80.40
& 29.04
& 43.70
& 38.76 \\

OPD+RL
& 22.60
& 23.96
& 18.65
& 9.22
& 12.08
& 66.88
& 84.61
& 81.60
& 30.51
& 44.44
& 39.45 \\

ExOPD
& 22.19
& 23.96
& 18.85
& 9.22
& 11.35
& 67.81
& 84.53
& 80.20
& 28.31
& 42.96
& 38.93 \\

\rowcolor{blue!5}
\textbf{Distilled RL}
& \textbf{24.38}
& \textbf{24.79}
& \textbf{19.17}
& \textbf{10.08}
& \textbf{12.40}
& \textbf{68.13}
& \textbf{85.82}
& \textbf{82.00}
& \textbf{31.62}
& \textbf{44.59}
& \textbf{40.30} \\

\bottomrule
\end{tabular}%
}
\end{table*}

\begin{table*}[t]
\caption{
Additional generalization and Pass@16 results on Qwen3-1.7B.
}
\label{tab:qwen17_generalization}
\centering
\resizebox{\textwidth}{!}{%
\renewcommand{\arraystretch}{1.30}
\begin{tabular}{
    >{\raggedright\arraybackslash}m{2.8cm}
    *{2}{>{\centering\arraybackslash}m{1.7cm}}
    *{1}{>{\centering\arraybackslash}m{2.1cm}}
    *{6}{>{\centering\arraybackslash}m{1.55cm}}
}
\toprule
\multirow{2}{*}{\textbf{Method}}
& \multicolumn{3}{c}{\textbf{Pass@1}}
& \multicolumn{6}{c}{\textbf{Pass@16}} \\
\cmidrule(lr){2-4}
\cmidrule(lr){5-10}

& \textbf{MMLU$_{\text{pro}}$}
& \textbf{SuperGPQA}
& \textbf{LiveCodeBench}
& \textbf{AIME24}
& \textbf{AIME25}
& \textbf{AIME26}
& \textbf{AMC23}
& \textbf{CMIMC25}
& \textbf{HMMT25} \\
\midrule

\rowcolor{gray!10}
Qwen3-1.7B
& 64.46
& 28.00
& 29.67
& 53.33
& 36.67
& 38.33
& 88.75
& 18.75
& 26.67 \\

OPD
& 63.92
& 29.20
& 34.12
& 68.33
& 46.67
& 50.00
& 92.50
& 32.50
& 33.33 \\

RL (GRPO)
& 65.71
& 31.00
& 36.78
& 63.33
& 46.67
& 48.33
& 92.50
& 31.25
& \textbf{35.00} \\

OPD+RL
& 65.17
& 30.60
& 33.36
& \textbf{70.00}
& 45.00
& 48.33
& 93.75
& \textbf{33.75}
& 31.67 \\

\rowcolor{blue!5}
\textbf{Distilled RL}
& \textbf{65.89}
& \textbf{31.60}
& \textbf{38.67}
& \textbf{70.00}
& \textbf{50.00}
& \textbf{51.67}
& \textbf{95.00}
& \textbf{33.75}
& \textbf{35.00} \\

\bottomrule
\end{tabular}%
}
\end{table*}

\subsection{Statistics of teacher--student importance ratios. \label{app:sub_rho}}
\begin{figure*}[t]
\centering
\includegraphics[width=1\textwidth]{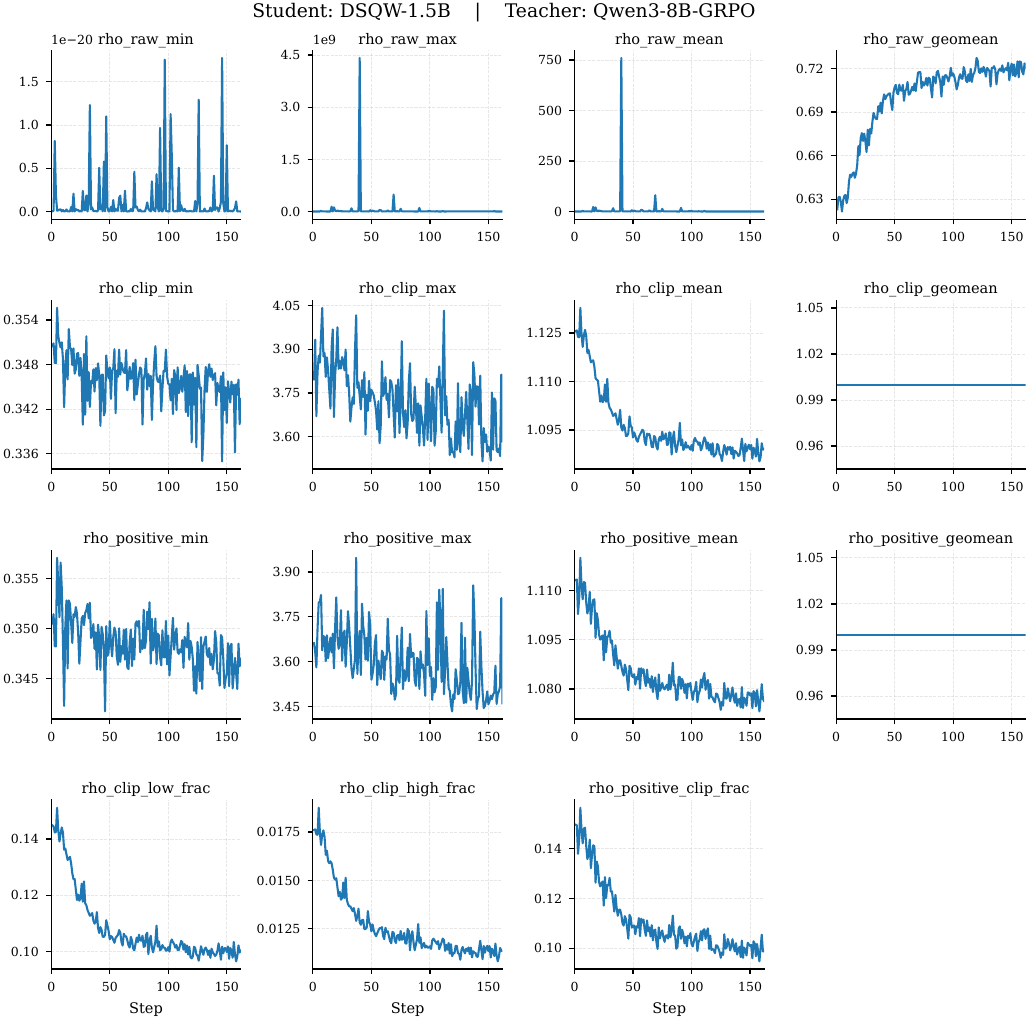}
\caption{
Statistics of the teacher--student importance ratio $\rho$ during Distilled RL training with DSQW-1.5B as the student and Qwen3-8B-GRPO as the teacher.
}
\label{fig:rho_dsqw}
\end{figure*}

\begin{figure*}[t]
\centering
\includegraphics[width=1\textwidth]{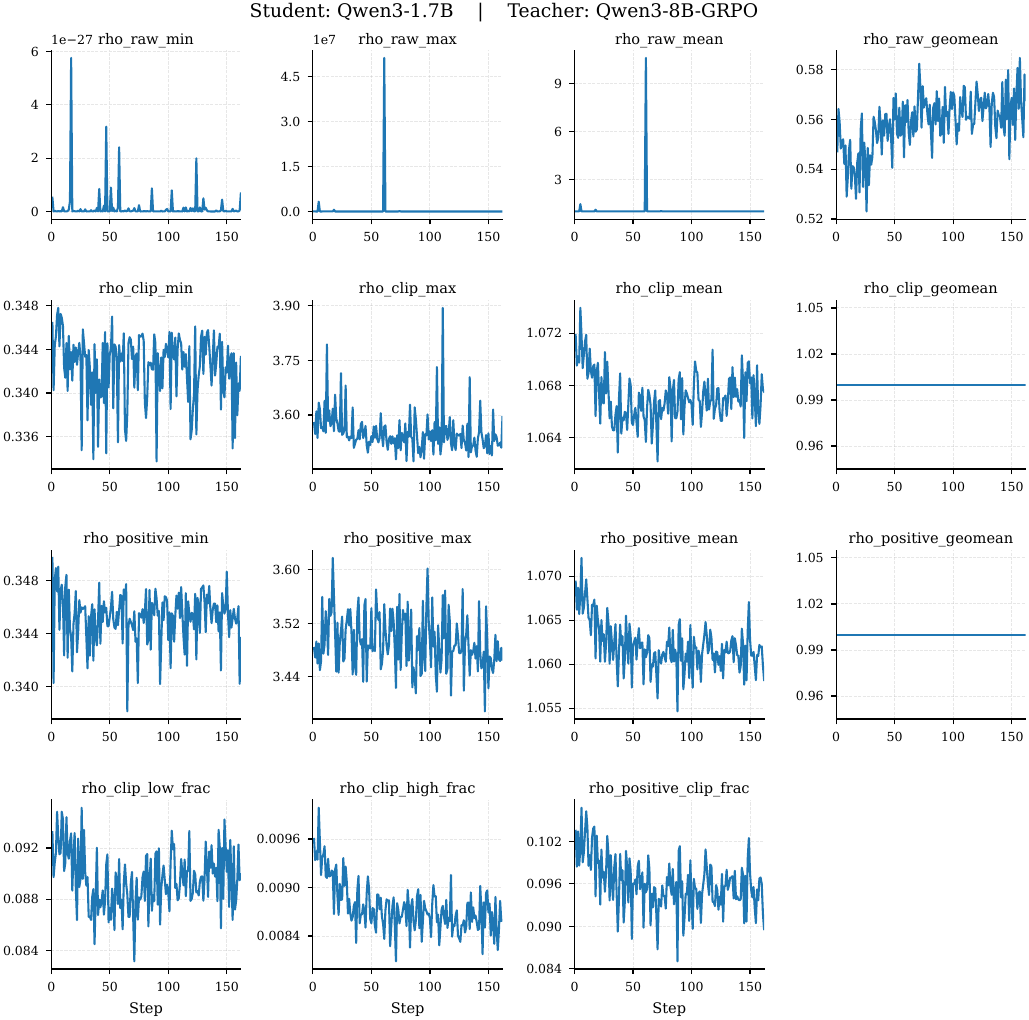}
\caption{
Statistics of the teacher--student importance ratio $\rho$ during Distilled RL training with Qwen3-1.7B as the student and Qwen3-8B-GRPO as the teacher.
}
\label{fig:rho_qwen17b}
\end{figure*}

\begin{figure*}[t]
\centering
\includegraphics[width=1\textwidth]{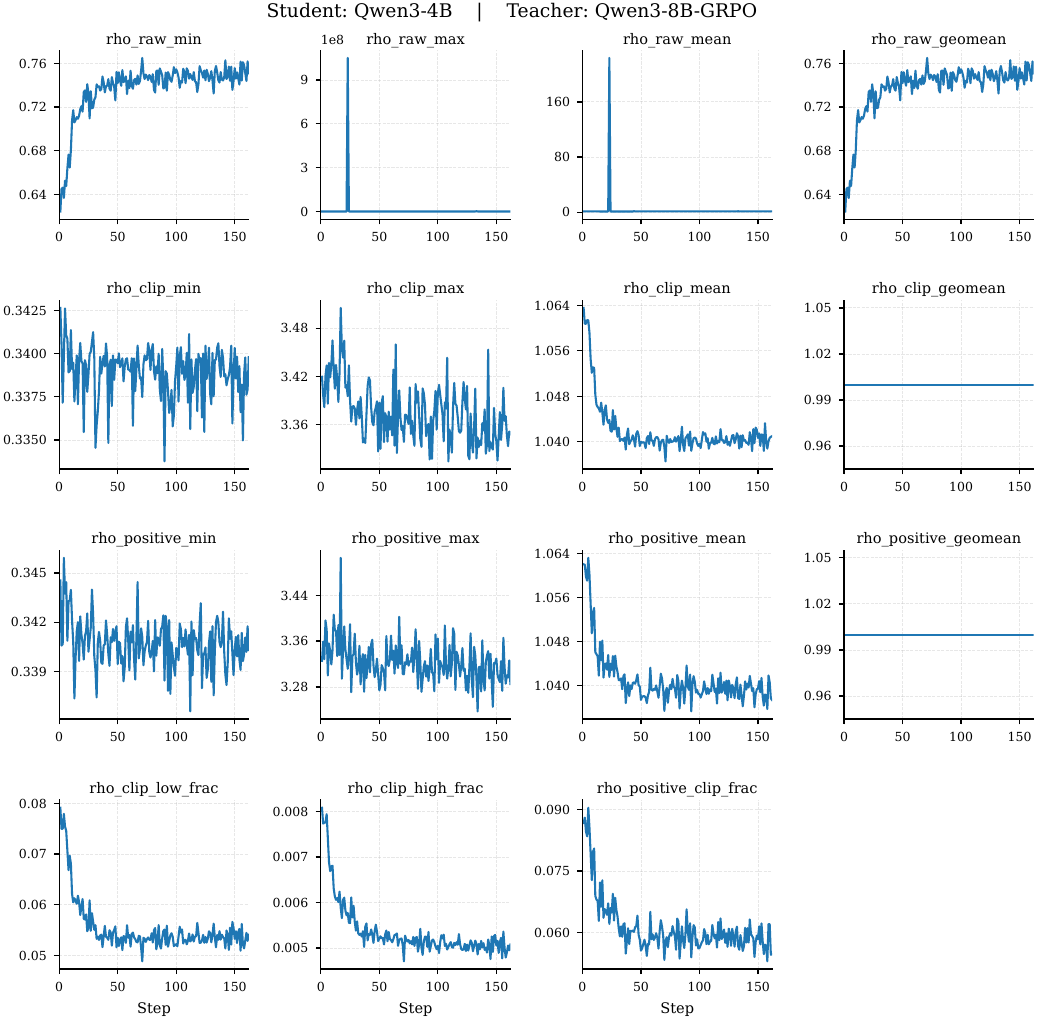}
\caption{
Statistics of the teacher--student importance ratio $\rho$ during Distilled RL training with Qwen3-4B as the student and Qwen3-8B-GRPO as the teacher.
}
\label{fig:rho_qwen4b}
\end{figure*}

To better understand the behavior of reverse importance sampling, we visualize the statistics of the teacher--student ratio $\rho$ throughout training in Figures~\ref{fig:rho_dsqw}--\ref{fig:rho_qwen4b}. Across all student models, the raw importance ratios exhibit a pronounced heavy-tailed distribution: the maximum and arithmetic mean can be dominated by occasional extreme values, whereas the geometric mean remains substantially more stable. This observation motivates explicitly clipping the importance ratios before using them for policy optimization.

After clipping, the ratio statistics become substantially more stable, showing that clipping effectively prevents a small number of extreme teacher--student likelihood ratios from dominating the update. The fraction of clipped tokens also generally decreases during training, particularly for DSQW-1.5B and Qwen3-4B, suggesting that the student distribution gradually becomes better aligned with the teacher. Notably, DSQW-1.5B exhibits a consistently larger clipping fraction than Qwen3-4B, consistent with the larger distributional mismatch in the cross-family distillation setting.

Finally, although the raw ratios tend to have a geometric mean below one, sequence-level geometric normalization keeps the effective geometric mean at one. This removes the systematic attenuation of positive trajectories while retaining the teacher's relative token-level preferences. Similar statistics are observed on positive-advantage samples, indicating that clipping and normalization remain necessary even after negative sample reset.

\section{Implementation Details}
\label{app:detail}

\begin{table}[t]
\centering
\caption{Detailed configurations in the experiments.}
\label{tab:implementation}
\renewcommand{\arraystretch}{1.12}
\setlength{\tabcolsep}{5pt}
\begin{tabular}{l l}
\hline
\textbf{Configuration} & \textbf{Setting} \\
\hline
\textbf{Training Settings} & \\
\quad Global batch size $|B|$ & 128 \\
\quad Mini-batch size & 64 \\
\quad Micro-batch size per GPU & 2 \\
\quad Rollout group size $G$ & 8 \\
\quad Maximum prompt length & 2048 \\
\quad Maximum response length & 8192 \\
\quad Sampling temperature & 1 \\
\quad Top-$p$ & 0.99 \\
\quad Learning rate & 1e-6 \\
\quad Number of training steps & 160 \\
\hline
\textbf{Policy Optimization} & \\
\quad Advantage estimation & Group-normalized outcome reward \\
\quad Reward type & Binary correctness reward \\
\quad Lower policy clip $\epsilon_{\mathrm{low}}$ & 0.2 \\
\quad Upper policy clip $\epsilon_{\mathrm{high}}$ & 0.2 \\
\hline
\textbf{Distilled RL Settings} & \\
\quad Teacher--student ratio &
$\rho_{i,t}=\pi_{\mathrm{teacher}}/\pi_{\theta_{\mathrm{old}}}$ \\
\quad Ratio clipping range &
$[1/\epsilon_{\rho},\,\epsilon_{\rho}]$ \\
\quad Ratio clipping threshold $\epsilon_{\rho}$ & 3 \\
\quad Negative sample reset &
$w_{i,t}=1$ when $A_i\leq 0$ \\
\quad Geometric normalization &
Sequence-level geometric mean normalization \\
\hline
\textbf{Baseline Settings} & \\
\quad RL & GRPO \\
\quad OPD divergence & Reverse KL divergence \\
\quad OPD+RL mixing coefficient & 1.0 \\
\hline
\textbf{Evaluation Settings} & \\
\quad Maximum response length & 8192 \\
\quad Temperature & 0.1 \\
\quad Top-$p$ & 0.95 \\
\quad Number of samples for Pass@1 & 32 \\
\quad Number of samples for Pass@16 & 32 \\
\hline
\end{tabular}
\end{table}

\paragraph{Experimental setup.}
We implement Distilled RL using the EasyR1 and VeRL frameworks \citep{zheng2025easyr1, sheng2025hybridflow}. Unless otherwise specified, all methods are trained on DAPO-17K \citep{yu2025dapo} using the same prompt distribution, rollout budget, optimization schedule, and evaluation protocol. We use Qwen3-8B-GRPO as the teacher model and evaluate two student models: Qwen3-4B and DeepSeek-R1-Distill-Qwen-1.5B, abbreviated as DSQW-1.5B \citep{yang2024qwen2, yang2025qwen3, guo2025deepseekr1}. Unless otherwise specified, the teacher model is trained on DAPO-17K under the same training setup for one epoch.
The complete training configuration is summarized in Table~\ref{tab:implementation}. 

For each training prompt $q$, the old student policy
$\pi_{\theta_{\mathrm{old}}}$ generates a group of $G=8$ responses. Each response is evaluated using a binary correctness reward. Following GRPO \citep{shao2024deepseekmath}, we normalize the rewards within each rollout group to obtain the response-level advantage.

\paragraph{Evaluation benchmarks.}
We evaluate mathematical reasoning on AIME24, AIME25, AIME26, CMIMC25, HMMT25, AMC23, GSM8K, MATH500, Minerva Math, and Olympiad \citep{balunovic2025matharena, lightman2023lets, cobbe2021gsm8k, lewkowycz2022solving}. The first five datasets are recent competition benchmarks and are also used to report the averaged competition score in the training-dynamics figures. The remaining datasets cover a broader range of grade-school, competition-level, and formal mathematical reasoning problems.

To examine whether post-training preserves or improves capabilities beyond the primary mathematical evaluation, we further assess generalization on MMLU-Pro \citep{wang2024mmlu}, SuperGPQA \citep{du2025supergpqa} and Livecodebench \cite{jain2025livecodebench}. Since the full SuperGPQA benchmark is relatively large and computationally expensive to evaluate, we construct a fixed subset of 500 examples for evaluation (selected by the shortest answer for verifiable reward). No examples from either benchmark are used as supervised training targets.

\end{document}